\RequirePackage{fix-cm}

\documentclass[smallextended]{svjour3}
\smartqed  

\usepackage{graphicx}
\usepackage{authblk}
\usepackage{graphicx}
\usepackage{bm}
\usepackage[left=2cm, right=2cm, top=2cm]{geometry}
\usepackage[utf8]{inputenc} 
\usepackage{amsmath}
\usepackage{amsfonts} 
\usepackage[english]{babel}
\usepackage{caption}
\usepackage{subcaption}
\usepackage{enumerate}
\usepackage{float}
\usepackage[linkcolor=blue,colorlinks=true]{hyperref}
\usepackage{multirow}
\usepackage{mathptmx}      
\usepackage{subcaption}
\usepackage{comment}

\begin{document}

\title{Deep Neural Network in Cusp Catastrophe Model}


\author{
  Ranadeep Daw
  \and
  Zhuoqiong He
}


\institute{Ranadeep Daw \at
             University of Missouri \\
             \email{ranadeepdaw@mail.missouri.edu}           
          \and
           He, Zhuoqiong \at
             University of Missouri \\
             \email{hezh@missouri.edu}
}

\date{\today}

\maketitle

\begin{abstract}
Catastrophe theory was originally proposed to study dynamical systems that exhibit sudden shifts in behavior arising from small changes in input. These models are generally nonlinear in nature and they can generate reasonable explanation behind the sudden jumps in data. Among the different catastrophe models, Cusp Catastrophe attracted the most attention due to its relatively simpler dynamics and rich domain of application. But because of its complex nature, the parameter space becomes non-convex and so it's very hard to solve for the data generating parameters. In this paper, we demonstrated how a Machine learning based model can be successfully trained to learn the dynamics of the Cusp catastrophe models, without ever really figuring out the original model parameters. Simulation studies and application on a few famous datasets are used to validate our approach. To our knowledge, this is the first paper of such kind where a neural network based approach has been applied in Catastrophe models.

\end{abstract}

\section{\label{sec1}Introduction}
Catastrophe theory was originally proposed by Ren\'e Thom in 1960 \cite{thom}, which was later popularized by Christopher Zeeman in 1970s \cite{zeeman}. It was proposed to analyze the scenarios encountered in various branches of science like economics, behavior, health, social science etc where the response variables are hardly continuous or linear. Instead the response variables sometimes show instances of arbitrary jumps with very small perturbation in the input features. Widely used linear models can not explain the behavior in these cases. There are other examples of more complex cases where the response variables can be multi-modal based on same input conditions; i.e there can be more than $1$ possible values of the response variable for the exact same input. These scenarios are very hard to analyse mathematically with the commonly used models. Catastrophe models are proposed to generate explanation in these scenarios. The most famous model among the catastrophe models is the Cusp Catastrophe model. Cusp catastrophe model has a relatively simpler dynamics compared to other catastrophe models and it can be used to explain discontinuous, nonlinear, unimodal or bimodal response variables.

Over the years, people have suggested different methods to solve for the generating parameters in the cusp catastrophe models. The most notable ones are the PolyCusp model by Guastello \cite{doi:10.1002/bs.3830270305} \cite{GUASTELLO198961}, SDECusp Model by Cobb et al \cite{doi:10.1002/bs.3830230511} \cite{COBB1980311} \cite{10.2307/2288534} and a recent probabilistic approach (hereby known as RegCusp model) developed by Chen and Chen \cite{chen2017cusp}. The PolyCusp model has been criticized by Alexander et al in \cite{polyprob}, where the authors demonstrated that the model can't distinguish between a Cusp catastrophe model and a linear model. SDECusp model takes a stochastic differential equation approach and can be used in cross-sectional data, but it also fails in other data types like binary or count data. As a statistical point of view, the RegCusp model is of more interest to us since it considers normally distributed error terms, which is used to maximize the likelihood of the response variable given the input features. But the main problem there is that because of the highly non-convex behavior of the parameter space, it's very hard to optimize the loss function (usually the negative log-likelihood of the response or MSE as shown in \cite{chen2017cusp}). Using various simulation studies, we've found that the optimization algorithms usually get stuck in local minimas. But one interesting finding in our simulation study was that the numerical value of the loss function was very close to the desired true value, which made us think that there might be other ways to duplicate the original models utilizing these local minimas, i.e there might be a way to train an algorithm to mimic the cusp catastrophe model without ever truly solving for the generating parameters. Motivated by this, our idea was to test if Deep learning models \cite{Goodfellow-et-al-2016} can be successfully trained to capture data from a Cusp Catastrophe model. Deep learning models have become very popular because of their ability to model complex scenarios, some notable examples including image processing \cite{EGMONTPETERSEN20022279}, speech recognition \cite{SECOMANDI20001201}, game playing \cite{44806} etc. Our contribution in this paper is to further establish their applicability by successfully demonstrating how the variation of regular deep models, namely Mixture Density Networks, can be used to train Cusp Catastrophe models. We have used the previously proposed data generation algorithms here and also proposed another novel data generation process that allows more than one possible value of the response variable for the same input features. We show that Mixture Density Networks (MDN) are powerful enough to fit the cusp catastrophe data, irrespective of the different kinds of data generating models.

The rest of the paper has the following structure. In section \ref{sec2}, we've given the brief description of Cusp catastrophe models, RegCusp model, Deep Feedforward network and Mixture Density networks. Section \ref{sec3} talks about our proposed model that can be generate bimodal response. Section \ref{sec4} contains results from simulation study and section \ref{sec5} contains the application on some famous datasets for Cusp Catastrophe model. Finally we discussed our findings and further studies we are planning in catastrophe theory.

\section{\label{sec2} Overview}

In \cite{zeeman}, the author gave the idea of a Cusp catastrophe model by explaining how the behavior of a dog changes when it is introduced to input conditions causing both rage and fear simultaneously. A dog, without the introduction of any of these input conditions, is usually inactive or stays in peace. But simultaneous introduction of rage and fear can lead the dog's behavior to either of the following two outcomes - it either attacks the person responsible for the input features, or it becomes defensive and flees the scenario out of fear. The difference in these two possible responses is very large and the dog chooses one over another behavior with very little change in the input. Hence we can observe rapid change in the behavior of dog with very little perturbation in the input. Further, the least likely behavior of the dog in the above condition is to stay neutral. So researchers tried to come up with a model that allows the response variable to have values in one of the two extreme surfaces along with making it extremely unlikely to have value in a neutral surface in the middle of the two extreme surfaces. These features can be very easily be explained in Cusp Catastrophe model.

Driven by the key idea that under some applied force, any physical system tries to achieve an equilibrium state, the general catastrophe models have the following structure:

\begin{align}
\frac{\delta y_t}{\delta t} = \frac{\delta V(y_t; \bm{\gamma})}{\delta y_t}
\end{align}

where $y_t$ is the position of the system under study at time t and $V$ is the potential function due to the applied force. $\gamma$ is the unknown parameter vector of the system and needs to estimated here using the data. The equilibrium point of this dynamic system is achieved when the right hand side equals $0$. The temporal behavior of $y$ is complicated, but the system always moves towards the equilibrium point where $y$ does not change over time anymore. We are interested in the behavior of the cusp catastrophe model when it achieves equilibrium. The potential function for the Cusp catastrophe model and the equilibrium surface are given below by the following equations:

\begin{align}
&V(y_t; \alpha, \beta) = \alpha y_t + \frac{\beta}{2} y_t^2 - \frac{1}{4}y_t^4 \\
&0 = \frac{\delta V(y_t; \alpha, \beta)}{\delta t} = \frac{\delta y_t}{\delta t} = \alpha y_t + \beta y_t - y_t^3  \label{eq:1}
\end{align}

Here $\alpha$ and $\beta$ are called the asymmetry parameter and the bifurcation parameter respectively. Since in equilibrium state, the response does not change with time anymore, we can drop the suffix $t$ from \eqref{eq:1}.

\begin{figure}[ht]
\centering
\includegraphics[width=10cm, height=8cm]{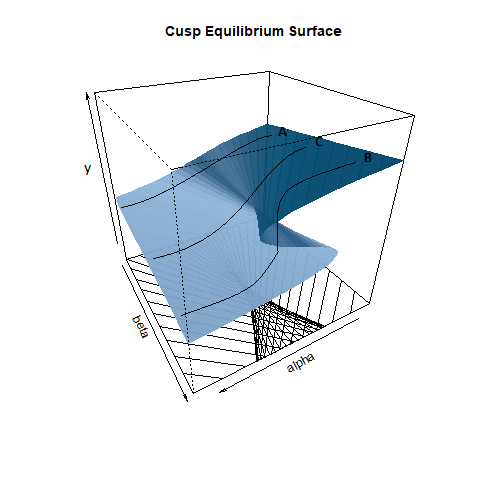}
\caption{Response surface of Cusp catastrophe model}
\label{cspresp}
\end{figure}

Figure \ref{cspresp} shows the response surface of the Cusp Catastrophe model. The response surface has a regular or stable behavior with respect to the control parameters in two regions, called the lower and upper stable regions. Away from this region, there can be $3$ typical behavior of $y$ that can be shown by lines {\textbf{A}-\textbf{C}}. Line \textbf{A} shows regular change in $y$ with change in $\alpha$ when the bifurcation parameter $\beta$ is smaller than a certain cutoff. When $\beta$ crosses a certain threshold, the typical behavior is like line \textbf{B}, where $y$ will have a sudden jump from the lower (or upper) stable region to the upper (or lower) one. This region looks like a cusp when projected to the ground or parameter space (the crossed area in the ground plane) and hence this parameter region is called the ``cusp region''. Line C is coming from outside the cusp region and hence even if the response surface goes through a sudden change from the upper to the lower stable region, there is no discontinuous jump here.

In the example of the behavior of the dog, the upper and lower stable regions can be thought of the 2 extreme behaviors (either attacking or defensive). The curved sheet like region is the most unlikely scenario when the parameters lie in the cusp region. This corresponds to the dog's neutral behavior when it faces rage and fear simultaneously, i.e when the dog is introduced to both rage and fear, the most unlikely behavior from the dog is to remain neutral. Since the equation \eqref{eq:1} can have $1$ or $3$ real roots depending on the values of $\alpha$ and $\beta$, these roots are leveraged to model the behaviors of the dog.

The model is completed with the introduction of independent feature variables. The feature variables are not directly modelled with the response. Instead they are modelled in a latent fashion by linking them to the two above control parameters $\alpha$ and $\beta$ by using a regression model. Mathematically, considering p feature variables $\{X_1, ... X_p\}$,  the control parameters are modelled as: 	

\begin{align}
\label{eq:control}
\begin{split}
&\alpha = a_0 + a_1 X_1 + ... + a_p X_p \\ 
&\beta = b_0 + b_1 X_1 + ... + b_p X_p   
\end{split}
\end{align}

We should further note that the number of roots of equation \eqref{eq:1} depends on Cardan discriminant (scaled) which is defined as: $\Delta = 27 \alpha^2 - 4 \beta^3 \label{eq:cardan}$. $\Delta$ $>$ $0$ implies only $1$ real root, whereas $\Delta$ $<$ $0$ implies 3 different real roots of equation \eqref{eq:1}. In the later case, there are two classical conventions to choose one real root, given by the following:

\begin{enumerate}[(i)]
\item \textbf{Delay convention} Choose the root closest to the observed data.
\item \textbf{Maxwell Convention} Choose the root that maximizes the highest potential value $V(y; \alpha, \beta)$. Following \cite{hartelman} and \cite{wag}, this also corresponds to the root that maximizes the likelihood of $y$ in equilibrium.
\end{enumerate} 

We should note that the above conventions prevents the model from having multiple responses for the exact same input condition. Without these conventions, the equation \eqref{eq:1} can have $1$, $2$ or $3$ roots, which means that the same input features can lead to $1$-$3$ different values for the response surface. This is hard to analyze in Mathematics since the traditional loss functions like squared error loss (SSE) or Mean Square Error (MSE) would be meaningless in this scenario. The Delay convention chooses the root that is closest to the observed response data, whereas Maxwell convention chooses the one that maximizes the likelihood of $y$. As already explained, in case of $3$ real roots, the root in the middle is most unlikely scenario. One of the other $2$ roots maximizes the potential and hence is selected in Maxwell convention. In case of only $1$ real root, we don't need to apply these conventions and the only real root is chosen as the solution of equation \eqref{eq:1}. 

The RegCusp model defined in \cite{chen2017cusp} follows the explained procedure to model the control parameters using equation \eqref{eq:control} and then uses the Maxwell convention to select the desired root $Y$. This root $Y$ is assumed to be the true value of the response surface, had there been no noise in the measurement. The observed output is then modeled by adding {\textit{iid}} Gaussian noises as the following: 
\begin{align}
    y_i = Y_i + \epsilon_i \text{\qquad $\epsilon_i$ $\stackrel{iid}{\sim}$ $\mathcal{N}(0, \sigma^2)$} \label{addnoise}
\end{align}

To solve for the model parameters, the authors tried to minimize the SSE or MSE. This in turn is equivalent to maximizing the conditional likelihood of $y$ given $Y$ and the MLE $\hat{\alpha}$, $\hat{\beta}$ and $\hat{\sigma^2}$. However, as already mentioned, the problem here is to optimize the loss function in the non-convex parameter space. Here we came up with the idea of using a Deep neural network to train the model.

Deep learning models are artificial neural networks (ANN) with one or more hidden layers. ANNs are computing systems inspired by the functionality of the neurons in our brains. These are collection of nodes (or neurons in Biology), where each node fires information to the nodes in the next layer only when it's value crosses some threshold, called the `activation' of the neuron. The most common example of an activation function here is ReLU that only keeps the positive part of a function like: $ReLU(x) = max(x, 0)$. A deep model can be achieved by adding neurons in multiple hidden layers. Hence each neuron can have $2$ states, namely active or firing information and inactive or not firing any information. These networks have been shown to be powerful enough to capture variety of complex dynamics. Mathematically, the $i$-th neuron in layer $l$ of a Deep network with $f$ as the particular activation function has the general form:
\begin{align}
    \mathbf{x}^{(l)}_i = f(\sum_j \mathbf{w}_{ij}^{(l-1)}\mathbf{x}_j^{(l-1)} + \mathbf{b}_i^{(l-1)})
\end{align}

where $\mathbf{w}_{ij}$ and $\mathbf{b}$ denotes the weight and the bias terms and $\mathbf{x}^{(0)}$ is the input data.

The predictions of a deep learning model does not contain any uncertainty quantification measurement. This can further be taken care of by using a mixture density layer as the last layer of the network. This kind of networks are known as Mixture Density Network (MDN). This was first proposed by Bishop in  \cite{mdn} to express neural networks as probabilistic models when they are given sufficiently enough amount of data. The primary assumption in an MDN is that any distribution can be arbitrarily approximated by a mixture of normal distributions. MDNs are recently becoming increasingly popular in this framework  (e.g., \cite{Davis551440}, \cite{817982}, \cite{makansi2019overcoming}). Mathematically, denoting $\langle Q \mid x \rangle$ = $\int_{\mathbf{t}} Q(\mathbf{t})p(\mathbf{t} \mid \mathbf{x})d\mathbf{t}$, the loss function (MSE) of a (deep) neural network can be approximated as:
\begin{align}
\begin{split}
    &L(\mathbf{w}; \mathbf{x}, \mathbf{y})\\
    &= \sum_{\mathbf{x}} \sum_{\mathbf{y}} [f(\mathbf{x}; \mathbf{w}) - \mathbf{y}]^2 \\
    & \approx \int_{\mathbf{x}} \int_{\mathbf{y}} [f(\mathbf{x}; \mathbf{w}) - \mathbf{y}]^2 p(\mathbf{y}, \mathbf{x}) d\mathbf{x} d\mathbf{y} \\
    &= \int_{\mathbf{x}} [f(\mathbf{x}; \mathbf{w}) - \langle \mathbf{y} \mid \mathbf{x} \rangle ]^2 p(\mathbf{x}) d\mathbf{x} + \\ 
    &\qquad \qquad \qquad \qquad 
    \int_{\mathbf{x}} [\langle \mathbf{y}^2 \mid \mathbf{x} \rangle - \langle \mathbf{y} \mid \mathbf{x} \rangle ^2] p(\mathbf{x}) d\mathbf{x} 
\end{split}
\label{eq:mdn}
\end{align}

The error of the network depends on the weights $\mathbf{w}$ through the first term only in the last line of equation \eqref{eq:mdn}. This shows that the fitted value obtained from a standard neural network is the conditional average of the response as a function of the input. This is true for any conditional distribution of $\mathbf{y} \mid \mathbf{x}$. So with the further assumption that any general distribution can be approximated by a Gaussian mixture density, $\mathbf{y} \mid \mathbf{x}$ is further modelled as a Gaussian mixture in the following equation:
\begin{align}
    p(\mathbf{y} \mid \mathbf{x}) = \sum_{i=1}^m \pi_i(\mathbf{x}) \phi(\frac{\mathbf{y} - \mu_i (\mathbf{x})}{\sigma_i(\mathbf{x})})
\end{align}

Here the constraints that $\pi_i$s are non-negative numbers that sum up to 1 and $\sigma_i$s are positive quantities are satisfied just like a Gaussian mixture model. The negative log-likelihood is taken as the loss function here, which means that the estimated response is obtained by maximizing the above likelihood function. Aggregating all these, an MDN can be written as  $(\bm{\mu}(\mathbf{x}), \bm{\sigma}(\mathbf{x}), \bm{\pi}(\mathbf{x}))$ where $\bm{\mu}$, $\bm{\sigma}$ and $\bm{\pi}$ are feed-forward deep networks with interpretations as mean, standard deviation and prior probabilities of the components and satisfy the constraints. Here we use the term $k$-components MDN to explain an MDN where the final layer is a $k$-component Gaussian mixture.

In our study, we initially generated data from RegCusp model and used an $1$-component MDN to predict the network. Since the data from RegCusp model is always unimodal due to the Maxwell convention, an $1$-component MDN turned out to be sufficient to predict the data, although the results improved with a $2$-component MDN. But this was not the case when we tried to fit some famous experimental cusp catastrophe datasets. These datasets rather show some evidences of multi-valued response variables; i.e same input features led to completely different response variables, roughly coming from the two different stable regions. Here a $1$-component MDN completely fails, but a $2$-component MDN turned out to be able to explain the data. Hence in the next section, we have recommended a slightly tweaked version of the RegCusp model for data generation to capture this bimodal nature. We call our model the `Bimodal RegCusp' model. We generated data from a variety of model in our study and demonstrated how deep networks can capture these data irrespective of the generating models.

\section{\label{sec3}Bimodal RegCusp model}

Our proposed Bimodal RegCusp model allows the response variable to lie in both the upper and lower stable regions when the parameters lie in the cusp region. When the parameter is outside the cusp region, there is only one possible real root of equation \eqref{eq:1} and so the problem of root selection does not appear there. Our model follows the same approach like RegCusp to link the input features and the latent control parameters using a linear regression like \eqref{eq:control}. Then the equation \ref{eq:1} is solved to obtain the true values of the response. To allow bimodality in the cusp region, we take the following approach. If there is only one real root, the root is considered as the true value of the response. In case of more than $1$ real root (i.e in the cusp region), we discard the unstable root (the middle one) and we choose one of the other two roots as the possible true values for the response surface with equal (or $0.5$) probability. The observed value of the response is then obtained by adding a Gaussian noise term like equation \eqref{addnoise}. 

\begin{figure}[ht]
\begin{subfigure}{.5\textwidth}
  \centering
  \includegraphics[width=.8\linewidth]{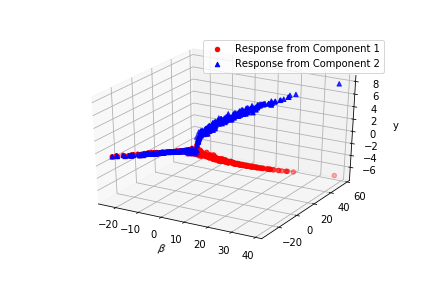}  
  \caption{Bimodal RegCusp Response surface}
\end{subfigure}
\begin{subfigure}{.5\textwidth}
  \centering
  \includegraphics[width=.8\linewidth]{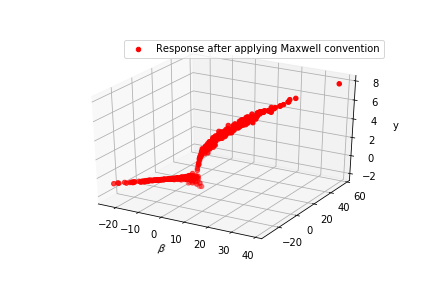}  
  \caption{RegCusp Response Surface}
 \end{subfigure}
\caption{Comparison of Bimodal RegCusp and RegCusp surface}
\label{fig:comparison}
\end{figure}

Figure \ref{fig:comparison} shows the difference in the response surfaces in RegCusp and Bimodal RegCusp model. In our approach, the region with small values of $\alpha$ and $\beta$ is unimodal or outside the cusp region and hence they have only one possible root. After a certain cutoff point, the cusp region starts and hence we get multiple roots as the solution of the equation \ref{eq:1}. The additional benefit here is that, since we are choosing one of the two possible roots with the same probability, both the upper and the lower stable regions are well represented here in the data. Our assumption here is that the response variable in the cusp region is actually bimodal and the two modes correspond to the two stable regions. This is not the same case like in Maxwell convention where only one of the stable surfaces can be represented by the equation. This is mathematically hard to analyse, but we demonstrate next how our approach has worked here.

As explained already, we used $1$-component MDN and $2$ -components MDN to fit the simulated datasets. From the theory of MDN, the last layer of a $k$-component MDN is a mixture of $k$ Gaussian distributions. Each of these distributions has a fitted mean $\hat{\mu_i}$, standard deviation $\hat{\sigma}$ and prior probability $\hat{\pi}$. So in $1$-component MDN, there is only one mean from the only Gaussian component. We choose this mean as the fitted response obtained from the fitted network. This is equivalent to the idea that the fitted mean is the MLE under the assumption of a normal distribution. This approach worked fine in case of RegCusp data, where the data is forced to be unimodal by applying Maxwell convention. But it failed completely in Bimodal RegCusp data and some famous experimental cusp datasets. So, we instead used a $2$-components MDN here. For the $2$-components MDN, each of the $2$ Gaussian component has a fitted mean and here our hypothesis is that each of these 2 means is used to fit one of the stable regions. Hence in the cusp region, the $2$-components MDN fits the bimodal nature of our response. This has a fundamental problem outside the cusp region because there are also $2$ different fitted values when there should be 1. But surprisingly we found that both the components of the $2$-components MDN almost overlap on each other outside the cusp region and hence the fitted values from the 2 components is almost close to one another. This convinced us that $2$-components MDN can be used here to fit the cusp catastrophe data.

The additional benefit here is that, even if we observe only one value of the response given the input, our network can predict where the other value of the response could have been given the same inputs. This was unanswered previously and now our approach has this novel feature. We further tried to demonstrate the goodness of fit of our model by calculating MSE using Delay convention. This means that to calculate MSE, we choose the mean that is closest to the observed response $y$ and use this mean as the fitted value of the response. Note that, the Delay convention here can never choose the unstable root since we are getting rid of that root before we apply the convention. In the next section, we demonstrated our simulation technique to generate data and then the model fitting technique is also discussed.

\section{\label{sec4}Simulation study}
In simulation study, we generated data from different cusp catastrophe models to show how powerful deep networks are. Irrespective of the generating model, MDNs can be trained successfully to learn the data generating process. The previously proposed methods usually fail here since one particular method usually does not perform too well with the data coming from a different model. Our success here is to show that our approach does not care for the generating model and they can be trained with very low computational cost.

\subsection{RegCusp}
We used 2 input features $X_1$ and $X_2$ as predictors. Following are the steps to generate the response surface:
\begin{enumerate}[i)] \label{gen}
\item Simulate n=500 independent variables $X_1$ and $X_2$ from normal distributions with mean $0$ and variance $4$.
\item The true regression coefficients; i.e, $a_0$, $a_1$, $a_2$, $b_0$, $b_1$, $b_2$ are generated uniformly from a regular grid between 0 to 5, $\sim$ $\mathcal{U}(0,5)$. 
\item The control parameters are calculated using \eqref{eq:control} as $\alpha$ = $a_0 + a_1 X_1 + a_2 X_2$, $\beta$ = $b_0 + b_1 X_1 + b_2 X_2$. Then solve the \eqref{eq:1} to get the roots.
\item Apply Maxwell convention to select one root that maximizes the potential function.
\item Simulate errors $\epsilon_i$ from $\mathcal{N}(0, 1)$ and produce the observed output $y_i$ = $Y_i$ + $\epsilon_i$.
\end{enumerate}

The data is randomly divided into train and test data ($50\%$ in each). For simulated data, we used MDN with $3$ hidden networks with dropout between layers. The predictor variables are standardized before passing to the network. We tried the same for the response also, but it didn't improve the result. We also tried batch-normalization layers which also was unnecessary. Different optimizers including sgd, rmsprop, adam etc produced similar results. We should note that the network might need slight tuning for each of the datasets, but overall these values of the hyperparameters are similar to each other. Different parameter settings are used and the MSE from these different simulations are reported in table \ref{simtable}.

\begin{table*}[ht]
\centering
\begin{tabular}{|r|r|r|r|r|r|r|r|}
\hline
$a_0$ & $a_1$ & $a_2$ & $b_0$ & $b_1$ & $b_2$ & MSE (1-Component) & MSE (2-component) \\
\hline
0.8374 & 0.5228 & 3.1822 & 3.5324 & 0.1579 & 4.6811 & 1.207 & 0.8773 \\
1.7122 & 3.8342 & 2.4415 & 2.7407 & 3.1888 & 4.0322 & 1.0242 & 0.9468 \\
1.198 & 2.7108 & 4.0073 & 2.1903 & 4.3106 & 4.5244 & 1.2273 & 0.9539 \\
0.419 & 0.6107 & 3.5677 & 1.8378 & 3.1572 & 3.4127 & 0.9218 & 1.08 \\
4.2665 & 2.6617 & 3.6516 & 0.8548 & 3.5857 & 4.0862 & 1.0409 & 1.0241 \\
\hline
\end{tabular}
\caption{MSE obtained from 1 and 2-component MDN}
\label{simtable}
\end{table*}

\begin{figure}[h]
     \centering
     \begin{subfigure}[b]{0.45\textwidth}
         \centering
         \includegraphics[width=\textwidth]{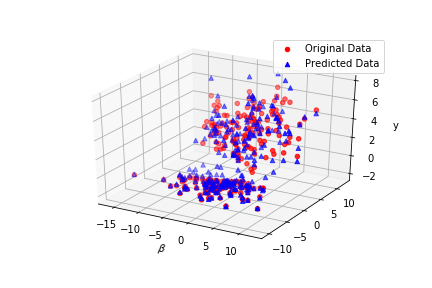}
         \caption{1-component MDN}
     \end{subfigure}
     \hfill
     \begin{subfigure}[b]{0.45\textwidth}
         \centering
         \includegraphics[width=\textwidth]{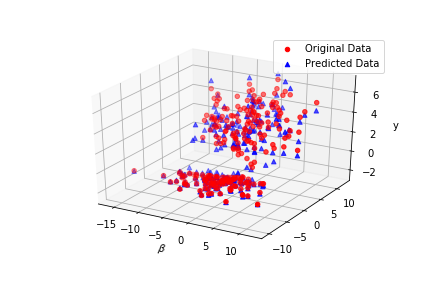}
         \caption{2-component MDN}
     \end{subfigure}
        \caption{RegCusp data fitting using MDN}
        \label{fig:gr1}
\end{figure}

Figure \ref{fig:gr1} shows the fit obtained in RegCusp data using $1$ and $2$-components MDN. Table \ref{simtable} shows the test MSE in $2$-component MDN has slightly better MSE compared to the $1$-component MDN. As mentioned earlier, we used Delay convention to calculate the MSE in case of 2-component MDN. The estimated MSE in all the cases are close to the true error variance, which shows that the variability in the data can be well-captured by MDN.

\subsection{Bimodal RegCusp}
To generate data from Bimodal RegCusp, we follow \ref{gen} upto step iv. Then, step v is replaced with the following:
\begin{enumerate}[i)]
\setcounter{enumi}{4}
\item If \eqref{eq:1} has only $1$ real root, select that one. If it has $3$ roots, discard the root in the middle. Then choose one of the other two roots with probability $0.5$ as the true value of the response.
\end{enumerate}

Again considering the roots as the true value of the response, we simulate Gaussian errors and add them to get the observed response (like step vi in \ref{gen}). As shown in figure \ref{fig:comparison}, both the upper and lower stable regions can be represented in Bimodal RegCusp model by the data here when the parameters lie in the cusp region. We applied both $1$-component and $2$-components MDN here to demonstrate why the $1$-component MDN falls short here. The following figure \eqref{fig:gr2} shows the fits obtained from both of these networks. Delay convention is also applied in the $2$-components MDN to show that the MSE is very low and hence this convinces us to use MDN in Bimodal RegCusp data.

\begin{figure}[ht]
     \centering
     \begin{subfigure}[b]{0.45\textwidth}
         \centering
         \includegraphics[width=\textwidth]{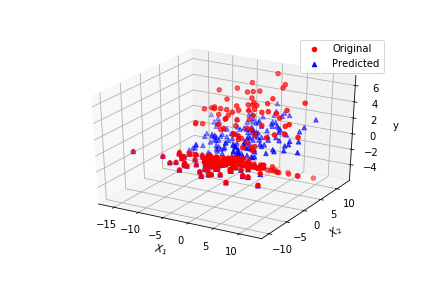}
		 \caption{1-component MDN}
     \end{subfigure}
     \hfill
     \begin{subfigure}[b]{0.45\textwidth}
         \centering
         \includegraphics[width=\textwidth]{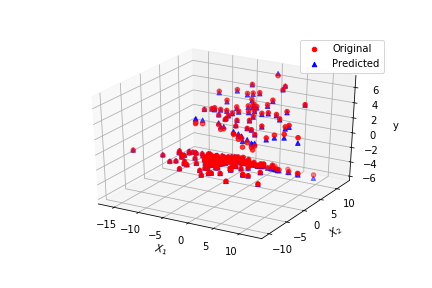}
		 \caption{2-component MDN}
     \end{subfigure}
        \caption{Bimodal RegCusp data fitting using MDN}
        \label{fig:gr2}
\end{figure}

In figure \ref{fig:gr2}, we can clearly see that in 1-component MDN, the response variable in the cusp region is not properly being fitted. Instead of fitting the two stable regions, the fitted network passes through the middle of the two possible response surfaces in the cusp region. Hence the network captures neither the upper nor lower stable response surfaces here. Instead of that, the $2$-components is fitting both the surfaces with its 2 means from the 2 components. The MSE in different datasets here turned out to be $7.86$ in $1$-component MDN, but that went down to $0.78$ in case of the $2$-component MDN. So this helped us to conclude that Bimodal RegCusp data needs a 2-components MDN to fit the data.

\subsection{SDECusp}
We also tried to use MDNs to fit data coming from SDECusp model proposed by Cobb. Here data is generated in \textbf{R} using the "cusp" package following \cite{rcusp}. To generate data, we followed algorithm \ref{gen} up to step (iii), where we generated $\alpha$ and $\beta$ values. Then we passed these values of the control parameters to the function \textit{rcusp} to generate data following SDECusp. As previously stated, we found that RegCusp method fails to estimate the parameters in SDECusp and vise versa. But using MDN, we can also fit the data with a very similar technique.

\begin{figure}[H]
     \centering
     \begin{subfigure}[b]{0.45\textwidth}
         \centering
         \includegraphics[width=\textwidth]{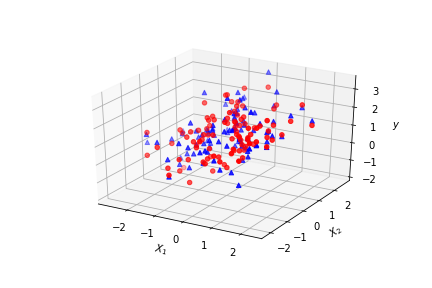}
		 \caption{1-component MDN}
     \end{subfigure}
     \hfill
     \begin{subfigure}[b]{0.45\textwidth}
         \centering
         \includegraphics[width=\textwidth]{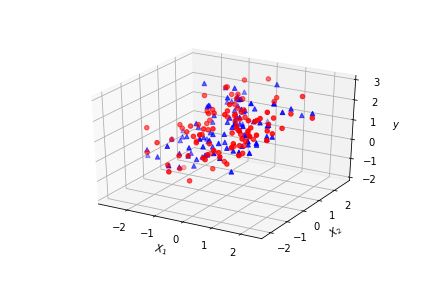}
		 \caption{2-component MDN}
     \end{subfigure}
     \caption{SDECusp Data fitting using MDN}
        \label{fig:RegCusp}
\end{figure}

Again we tried both $1$-component MDN and $2$-component MDN here. Figure \ref{fig:RegCusp} shows the fit in the test data after training is done. Our fit shows that there is not much difference between the results obtained from 1-component and 2-component MDN here. 

\section{\label{sec5}Application on famous Datasets}

\subsection{Zeeman datasets}
There are a few experimental datasets in R available in the "cusp" package, including data from Zeeman's catastrophe machine, namely $zeeman1$, $zeeman2$ and $zeeman3$. We used both 1-component and 2-component MDNs here to fit the data. Again Delay convention is used in $2$-component MDN. We found that in all 3 datasets, the result improves in the $2$-component MDN. But the interesting fact here is that the fit in $zeeman1$ and $zeeman2$ datasets from an $1$-component MDN is reasonable enough, but this completely fails in $zeeman3$. Only a $2$-component MDN can fit $zeeman3$. Our approach improves the MSE in all 3 datasets compared to the MSE obtained from the cusp function from \textbf{R}, explained in \cite{rcusp}. Further, the cusp package uses the whole data to fit the network, where as our approach improves MSE even after using only part of the data to fit the networks. Table \ref{msetable} provides the comparisons between MSEs.

\begin{table}[ht]
\centering
\begin{tabular}{|r|rrr|}
\hline
\multirow{2}{*}{Dataset} & \multicolumn{3}{c|}{MSE} \\
& 1-component MDN & 2-components MDN & cusp package \\
\hline
$zeeman1$ & 0.14 & 0.10 & 0.19 \\
$zeeman2$ & 0.25 & 0.20 & 0.21 \\
$zeeman3$ & 7.86 & 0.79 & 0.88 \\
\hline
\end{tabular}
\caption{MSE in $zeeman$ datasets from MDN}
\label{msetable}
\end{table}

\begin{figure}[H]
     \centering
    \begin{subfigure}[b]{0.4\textwidth}	
         \centering
         \includegraphics[width=\textwidth, height=6cm]{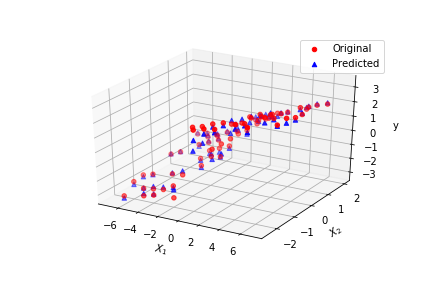}
		 \subcaption{Test Data for $zeeman1$}
     \end{subfigure}
     \begin{subfigure}[b]{0.4\textwidth}
         \centering
         \includegraphics[width=\textwidth, height=6cm]{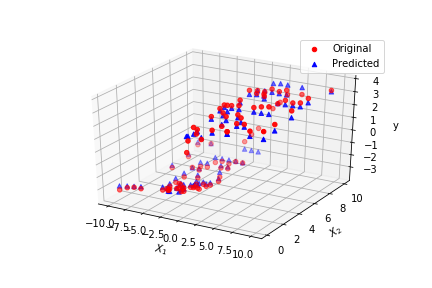}
		 \subcaption{Test Data for $zeeman2$}
     \end{subfigure}
     \caption{Fits using 2 component MDN}
     \label{z1z2}
\end{figure}

In figure \ref{z1z2}, we have demonstrated the fit in $zeeman1$ and $zeeman2$ from the 2-components MDN. The fit from $1$-component MDN is very similar to the results from $2$-component MDN here.

\begin{figure}[H]
     \centering
     \begin{subfigure}[b]{0.4\textwidth}	
         \centering
         \includegraphics[width=\textwidth, height=6cm]{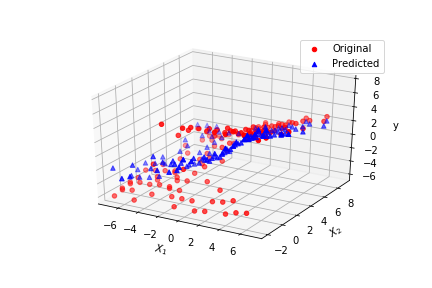}
		 \subcaption{$zeeman3$ fit from 1-component MDN}
     \end{subfigure}
     \begin{subfigure}[b]{0.4\textwidth}	
         \centering
         \includegraphics[width=\textwidth, height=6cm]{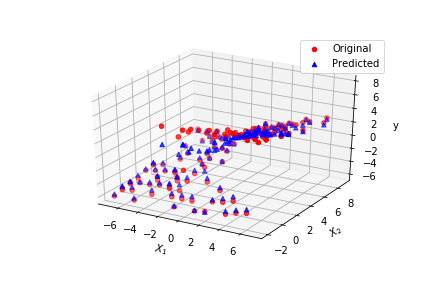}
		 \subcaption{$zeeman3$ fit from 2-component MDN}
     \end{subfigure}
     \caption{Fit $zeeman3$ using 1 and 2-components MDN}
        \label{fig:z3better}
\end{figure}

Figure \ref{fig:z3better} shows that the 1-component MDN fails completely here to capture the response surfaces. Here the 2-component MDN shows its worth by successfully capturing both the stable regions.

\subsection{Oliva data}
Oliva data is also obtained from the cusp package in R. It was described in \cite{doi:10.1002/bs.3830320205} for experimental analysis of the cusp models. The data generation process there was slightly different from our considered models. They can be generated using the following equations:

\begin{align}
\begin{split}
&\alpha_i = X_{i1} - .969\,X_{i2} - .201\,X_{i3} \\
&\beta_i = .44\,Y_{i1} + 0.08\,Y_{i2} + .67\,Y_{i3} + .19\,Y_{i4} \\
&Z_i = -0.52\,U_{i1} - 1.60\,U_{i2} \\
\end{split}
\end{align}

Here $X_{ij}$ is uniformly distributed on $(-2,2)$, and $Y_{ij}$ and $U_{i1}$ are uniform on $(-3,3)$. The states $Z_i$ were then generated from the cusp density, using rcusp, with their respective $\alpha_i$ and $\beta_i$ as normal and splitting factors, and then $U_2$ was computed as:
\begin{align}
    U_{i2} = \frac{(Z_i + 0.52 U_{i1} )}{1.60}
\end{align}

We considered $Z$ as response and $x_1$, $x_2$, $x_3$, $y_1$, $y_2$, $y_3$, $y_4$ as predictors. We used both $1$-component and $2$-component MDNs to compare the fits in the test data. Following is the fit from $1$ and $2$ components MDN in the Oliva data.

\begin{figure}[H]
     \centering
    \begin{subfigure}[b]{0.4\textwidth}	
         \centering
         \includegraphics[width=\textwidth, height=6cm]{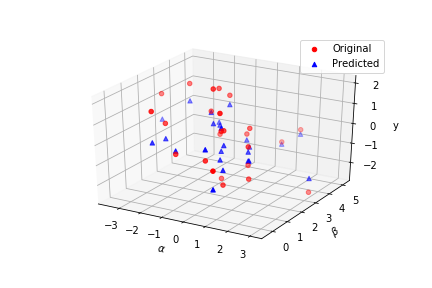}
		 \subcaption{Fit using 1-component MDN}
     \end{subfigure}
     \hfill
     \begin{subfigure}[b]{0.4\textwidth}	
         \centering
         \includegraphics[width=\textwidth, height=6cm]{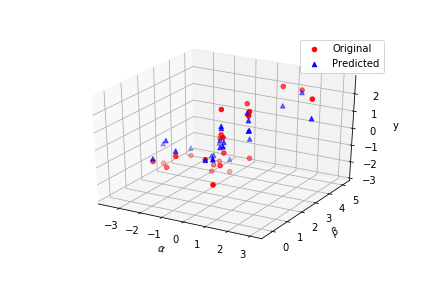}
		 \subcaption{Fit using 2-components MDN}
     \end{subfigure}
     \caption{Oliva test data}
\end{figure}

The MSE in $1$-component MDN is 1.12, whereas it is 0.73 in the $2$-component MDN. We found that the cusp package in R produces lower MSE than our approach. But the interesting thing here is that when we used 50\% data to fit the model in R, the test MSE was $9.88$ which is much higher compared to our approach. Also, we should note that Oliva is a synthetic data that was generated following the SDECusp model and the cusp package is also developed specifically to solve data SDECusp data. Hence `cusp' package fits this data better when we used the full data. Instead, our approach fits data from all cusp catastrophe models reasonably well.

\section{\label{sec6}Conclusion}
In this paper, we have demonstrated how deep neural networks and specifically Mixture Density Networks (MDN) can be used to fit Cusp Catastrophe model. We also took a novel model of generating data that can take care of both the stable regions when the parameters are in cusp region. Our approach has been tested on simulated data from different types of simulation methods. Also some famous datasets were used too to see if the results improve through our approach.

Our main goal here has been to show how effective the recently developed machine learning structures can be in the study of catastrophe models. An $1$-component MDN turned out to be good enough for RegCusp like data, whereas real life datasets like $zeeman3$ or oliva performed better with a 2-component MDN. A $2$-component MDN can even further predict the response in both the upper and lower response surfaces for each input, even though we only observe data from one of the response surfaces for each input data. Then a Delay convention is used here to compute MSE. These deep models are very fast to train and shows promise to be effective in predicting the response.  Our study further demonstrates how powerful deep learning models are and how the deep learning models can be equipped with statistical interpretation by adding mild assumption on the residuals.

There is one thing worth mentioning here. MDN fits the conditional distribution of the response given the inputs. With further assumption that the conditional distribution is mixture normal, we used the means of the Gaussian components to represent the two possible values in the cusp region or the two means overlap almost one another in the non-cusp region. Instead of this, we could have taken the approach of generating the response from the predicted distribution (Gaussian Mixture) obtained from the network. We also tried this approach and the results were pretty similar to the approach that we already took. Only difference we found is that, when we generate the response from the distribution, there is no way of ensuring that the generated distribution comes from the same stable region as the observed data in cusp region. The predicted value from the MDN in cusp region can lie in either of the two stable regions. Hence computing MSE becomes difficult here.

This opens up further possibilities of application. One straight forward advancement is to consider other types of response variables - like binary, count or longitudinal data. Also this can be applied on data from other catastrophe models. Also, another interesting application would be to come up with an algorithm that can conclusively capture the cusp region in the parameter space by using the input features. In conclusion, this paper broadens the application of machine learning models in behavioral science studies and opens up lots of possibilities for future work.

\section{Conflict of Interest}
On behalf of all authors, the corresponding author states that there is no conflict of interest.


%
%

\bibliographystyle{unsrt}
\bibliography{references}   

\end{document}